\PassOptionsToPackage{dvipsnames}{xcolor}
\documentclass[sigconf, nonacm, screen]{acmart}

\usepackage[utf8]{inputenc}
\usepackage[T1]{fontenc}
\usepackage{subcaption}
\usepackage{mathtools}
\usepackage{balance}
\usepackage{enumitem}
\usepackage{stfloats}
\usepackage{algpseudocode}
\usepackage{amsmath}
\usepackage[frozencache,cachedir=_minted]{minted}

\usepackage{booktabs}
\usepackage{color, colortbl}
\usepackage[ruled, linesnumbered, vlined]{algorithm2e}
\PassOptionsToPackage{hyphens}{url}
\usepackage{adjustbox}
\usepackage{microtype}
\usepackage{listings}
\usepackage{svg}
\usepackage{float}
\usepackage{multirow}
\usepackage{graphicx}
\usepackage
[
    noabbrev,   %
    nameinlink, %
]{cleveref}

\usepackage[normalem]{ulem}
\useunder{\uline}{\ul}{}

\usepackage{csquotes}
\usepackage{meta/dynamicwidth}
\usepackage{meta/circledsteps}

\usepackage{import}

\usepackage[most]{tcolorbox} %

\input{meta/commands}

\newcommand\vldbdoi{XX.XX/XXX.XX}
\newcommand\vldbpages{XXX-XXX}
\newcommand\vldbvolume{20}
\newcommand\vldbissue{1}
\newcommand\vldbyear{2027}
\newcommand\vldbauthors{\authors}
\newcommand\vldbtitle{\shorttitle}
\newcommand\vldbavailabilityurl{}
\newcommand\vldbpagestyle{plain}

\begin{document}

\title{\zephon: Elastic Determinism for Online, Stateful\\ Foundation Model Data Loading Pipelines}

\author{Maximilian Böther}
\affiliation{%
  \institution{DatologyAI \& ETH Zürich}
}
\authornote{Correspondence to \texttt{mboether@datologyai.com}}

\author{Josh Wills}
\affiliation{%
  \institution{DatologyAI}
}

\author{Ties Robroek}
\affiliation{%
  \institution{DatologyAI \& ITU Copenhagen}
}

\author{Sonnet Xu}
\affiliation{%
  \institution{Stanford University}
}
\authornote{Work done while at DatologyAI.}

\author{Paul Burstein}
\affiliation{%
  \institution{DatologyAI}
}

\author{Daniel Zayas}
\affiliation{%
  \institution{DatologyAI}
}

\author{Cody Blakeney}
\affiliation{%
  \institution{Arcee AI}
}
\authornotemark[2]

\author{Siddharth Joshi}
\affiliation{%
  \institution{DatologyAI}
}

\author{Haoli Yin}
\affiliation{%
  \institution{DatologyAI}
}

\author{Rishabh Adiga}
\affiliation{%
  \institution{DatologyAI}
}

\author{Haakon Mongstad}
\affiliation{%
  \institution{DatologyAI}
}

\author{Luke Merrick}
\affiliation{%
  \institution{DatologyAI}
}

\author{Pratyush Maini}
\affiliation{%
  \institution{DatologyAI \& CMU}
}

\author{Ari Morcos}
\affiliation{%
  \institution{DatologyAI}
}

\author{Matthew Leavitt}
\affiliation{%
  \institution{DatologyAI}
}

\author{Ana Klimovic}
\affiliation{%
  \institution{ETH Zürich}
}

\author{Bogdan Gaza}
\affiliation{%
  \institution{DatologyAI}
}

\begin{abstract}
Deterministic data loading is important for foundation model development: model researchers need confidence that differences they observe across costly ablations are caused by the parameter they changed rather than non-determinism in the training data sequence. 
The data loader must provide elastic determinism, i.e., a deterministic sequence of global training data batches despite changes to the GPU topology across runs (e.g., due to GPU scarcity), frequent checkpoint-resume cycles, and different data processing execution backends. 
Achieving this is difficult because modern foundation model data pipelines tokenize, pack, and mix samples online, introducing stateful $n$-to-$m$ transformations that break sample indexing.
Existing data loaders largely assume indexable $1$-to-$1$ pipelines, and the common workaround of offline materialization is expensive and, for some modalities such as video, infeasible.

We present \zephon, a data loader for foundation models that supports online, stateful pipelines while providing elastic determinism and efficient resumption from checkpoints.
It partitions the global stream into topology-independent lanes, serializes ordering decisions while parallelizing stateless work on interchangeable backends, and checkpoints only bounded in-flight state so recovery cost does not grow with training progress.
\revi{We evaluate \zephon{} on text and vision-language workloads and show that it achieves competitive throughput while providing a combination of guarantees that no existing loader offers for online, stateful pipelines.}

\end{abstract}

\maketitle

\pagestyle{\vldbpagestyle}
\begingroup\small\noindent\raggedright\textbf{PVLDB Reference Format:}\\
\vldbauthors. \vldbtitle. PVLDB, \vldbvolume(\vldbissue): \vldbpages, \vldbyear.\\
\href{https://doi.org/\vldbdoi}{doi:\vldbdoi}
\endgroup
\begingroup
\renewcommand\thefootnote{}\footnote{\noindent
This work is licensed under the Creative Commons BY-NC-ND 4.0 International License. Visit \url{https://creativecommons.org/licenses/by-nc-nd/4.0/} to view a copy of this license. For any use beyond those covered by this license, obtain permission by emailing \href{mailto:info@vldb.org}{info@vldb.org}. Copyright is held by the owner/author(s). Publication rights licensed to the VLDB Endowment. \\
\raggedright Proceedings of the VLDB Endowment, Vol. \vldbvolume, No. \vldbissue\ %
ISSN 2150-8097. \\
\href{https://doi.org/\vldbdoi}{doi:\vldbdoi} \\
}\addtocounter{footnote}{-1}\endgroup

\ifdefempty{\vldbavailabilityurl}{}{
\vspace{.3cm}
\begingroup\small\noindent\raggedright\textbf{PVLDB Artifact Availability:}\\
The source code, data, and/or other artifacts have been made available at \url{\vldbavailabilityurl}.
\endgroup
}

\section{Introduction}

\begin{figure}
  \begin{subfigure}{0.534\linewidth}
    \includeinkscape[width=\linewidth]{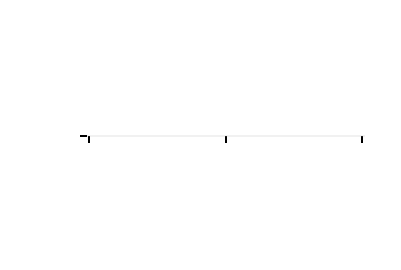}
    \captionsetup{margin={18pt,0pt}}
    \caption{Standard data loader}
    \label{fig:intro-el-nondet}
    \end{subfigure}%
    \begin{subfigure}{0.466\linewidth}
      \includeinkscape[width=\linewidth]{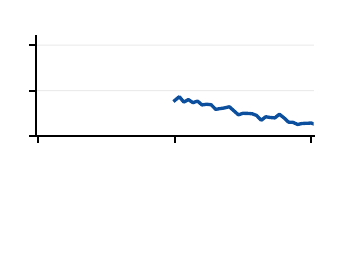}
      \caption{\zephon}
      \label{fig:intro-el-det}
    \end{subfigure}
    \caption{\revMRa{The effects of elastic determinism: a 1\,B text model is trained on 16 GPUs, checkpointed at step 150, and resumed on 8, 32, and 64 GPUs while the original run continues uninterrupted, causing loss variance between the runs when not using \zephon. The barcode below each plot shows the per-step batch SHA-256 hash of the 8-GPU continuation: blue shows matching batches while red shows differing batches.}}
    \label{fig:intro-el}
    \Description{TODO!}
    \vspace{-0.5cm}
  \end{figure}

A fundamental principle in experimental research is isolating variables, i.e., \emph{changing one thing at a time}, as stressed by early systems work such as~\citet{Saltzer1969Multics}.
If two experiments differ only in a single parameter while the rest of the system behaves deterministically, any change in outcome is attributable to that parameter alone.

Deterministic infrastructure is especially important for modern machine learning (ML) systems.
During foundation model development, practitioners run a
series of \emph{ablations} (prototype runs designed to isolate the effect of individual interventions, e.g., changing the data mix) before using the best-performing configuration for the large-scale \emph{hero run}~\cite{Groeneveld2024OLMo,Hoffmann2022ScalingLaws,Leavitt2022MosaicResNet,Gadre2025Scaling}.
Ablations, however, are costly, even at smaller scales.
\revi{For example, consider a 400\,B-total/20\,B-active Mixture-of-Experts hero run trained on 20\,T tokens. Even an ablation 1.5 orders of magnitude smaller in total training compute requires roughly 30\,k to 50\,k H100 GPU-hours, costing approximately \$250\,k at early-2026 cloud prices.}

Achieving reproducible ablations requires deterministic training infrastructure~\cite{Zhuang2022Randomness}.
One part of this infrastructure is the last-mile \emph{data loading}.
The data loader transforms an encoded representation of training data to tensors that the training framework passes to the model~\cite{Mohan2021DataStalls,Zhao2022MetaRM,Murray2021tfdata,graur2022cachew,graur2024pecan,Robroek2026TensorSocket,Kuchnik2022Plumber,Isenko2022Bottleneck}.
Without a deterministic data loader, the data ordering changes. 
This is problematic as the training loss, the common proxy for success of an ablation~\cite{Kaplan2020ScalingLaws,Hoffmann2022ScalingLaws,Gadre2025Scaling}, is sensitive to such changes in data order~\cite{Wang2025DataOrder,Biderman2023Pythia}.
\emph{Any} variation in the input data sequence confounds the results of ablations and impedes data debugging.
When a loss spike occurs, engineers must be able to inspect the offending sample(s).

Determinism with respect to a fixed cluster configuration is not sufficient.
In ML clusters, GPU resources are scarce. 
It is standard practice to prototype runs intended for large GPU counts on smaller allocations, using gradient accumulation to simulate the target global batch size~\cite{MosaicML2022Streaming,Wan2025ByteCheckpoint,AWS2026SageMakerElastic,Li2023EasyScale,Kang2025ElasWave}.
Moreover, cluster priorities may shift throughout the life of a training job.
A more urgent job may preempt part of an allocation, forcing a run to resume on fewer nodes. 
Conversely, when additional nodes become available, a job should be able to scale up to finish sooner.
In all these cases, the data loader must produce identical global training batches regardless of the GPU topology.
\revMRf{\textsc{Mosaic Streaming} calls this property \emph{elastic determinism}~\cite{MosaicMLStreamingElasticDeterminism}.}
\revMRa{\Cref{fig:intro-el} illustrates how, without elastic determinism, the loss curve varies when a training run is resumed on a different GPU count after a fault.
We use a standard 1\,B training setup in \textsc{Torchtitan} (sequence length 4\,096, AdamW, peak LR \(7.908 \cdot 10^{-4}\) with a WSD schedule, local batch size 7, global batch size 448), and use a shared initial seed checkpoint.
With \zephon, every topology receives the same global batches, confirmed by identical SHA hashes, causing the loss curves to overlap.
}

Beyond elastic determinism, the data loader must also be able to resume efficiently from a checkpoint~\cite{MosaicML2022StreamingResumption,Mohan2021CheckFreq,Wan2025ByteCheckpoint,Wan2025ByteRobust,Wang2023Gemini,Strati2025PCcheck}.
Resumptions are frequent and arise from both expected causes (e.g., preemption by higher-priority jobs or planned maintenance) and unexpected ones, such as hardware failures.
At scale, hardware failures are common, e.g., \citet{Kokolis2025Reliablity} project a mean-time-to-failure of 1.8\,h for 16\,k GPU training jobs.
As data loader state is part of every checkpoint~\cite{Wan2025ByteCheckpoint}, if restoring it requires replaying the full pipeline from the start, recovery time grows with training progress, and expensive GPUs sit idle waiting for data~\cite{Mohan2021DataStalls}.

However, implementing elastic determinism and efficient resumption for modern foundation model input data pipelines is hard because they consist of stateful operations with $n$-to-$m$ characteristics~\cite{An2025LLaVAOneVision15,Wang2025OpenQwen2VL,Zhao2025OVERLORD,Chen2024DataJuicer}.
\revMRf{In pre-training, tokenization splits documents into multiple subsequences~\cite{Sennrich2016BPE,Kudo2018SentencePiece}, packing maintains in-flight bins to fill fixed-length slots~\cite{Krell2021Packing,Zhao2024SequenceComposition}, and data mixing reorders samples~\cite{Zhao2025OVERLORD,Bother2026Mixtera}.
Post-training shares these characteristics: supervised fine-tuning packs chat-templated conversations with aligned response-only loss masks, while online reinforcement learning filters and regroups variable-cardinality rollouts.}
Whereas in a $1$-to-$1$ pipeline, progress is a sample index and resuming means looking up the last accessed sample, in a stateful $n$-to-$m$ pipeline, logical progress is spread across in-flight operator state, partially formed outputs, and reordered records.
Such pipelines are thus not \emph{indexable}.
Restoring the exact state without fully replaying all data and ensuring it remains identical across changes in parallelism, execution backend, or cluster topology is the core challenge.

\revi{Existing data loaders such as \textsc{tf.data}~\cite{Murray2021tfdata} or \textsc{Cedar}~\cite{Zhao2024cedar} optimize pipelines dominated by stateless $1$-to-$1$ mappings (e.g., decode a JPEG, flip an image).}
\revMRf{\textsc{Mosaic Streaming} provides elastic determinism for such indexable datasets.
\textsc{Grain} supports deterministic resumption for stateful pipelines under fixed execution configurations and elastic determinism for indexable pipelines~\cite{Grain2023Github}.}
While stream processing engines such as \textsc{Flink}~\cite{Carbone2015Flink} or \textsc{Naiad}~\cite{Murray2013Naiad} support stateful pipelines, their correctness model targets properties such as exactly-once processing, not a canonical output sequence of training batches that is preserved across topology changes.
\revi{To our knowledge, no existing system combines support for stateful $n$-to-$m$ pipelines with elastic determinism and efficient resumption.}

The common workaround is to materialize datasets in fully preprocessed form, i.e., tokenized, mixed, and packed into ready-made training batches~\cite{Shoeybi2019Megatron,NVIDIA2026MegatronData,Bother2026Mixtera,An2025LLaVAOneVision15,Wang2025OpenQwen2VL,Chen2024InternVL25,Zhu2025InternVL3,NVIDIA2024NeMoPacking}, using offline systems like \textsc{Spark}~\cite{Zaharia2016Spark}, \textsc{Beam}~\cite{Akidau2015Dataflow}, or \textsc{Ray}~\cite{Moritz2018Ray}, reducing the pipeline to an indexable $1$-to-$1$ mapping.
But model developers make frequent changes to the data pipeline (e.g., test a new filter), and re-materializing petabyte-scale corpora in offline systems on every change is \emph{prohibitively expensive in both storage and compute~\cite{Bother2026Mixtera}}.
For some data modalities, such as video, even materializing raw tensors is \emph{fundamentally not viable} due to the massive inflation in size~\cite{Zhao2025OVERLORD,vonWerra2026TRL}.
Labs that work with such modalities resort to building proprietary loading infrastructure~\cite{Anyscale2024Runway,Crosley2026VideoInfra,PerceptronAI2025Isaac}.
More broadly, the lack of a system that meets these requirements forces model teams to repeatedly reinvent data loaders, burning time and resources that could be better spent on training models and doing research.

The ML community needs an online data loading system that supports stateful $n$-to-$m$ operators while providing elastic determinism and efficient resumption.
Since data loading pipelines often consist of compute-intensive operators, they can bottleneck training and leave expensive GPUs idle waiting for data~\cite{Mohan2021DataStalls}. 
Decoupling data processing semantics from the execution backend is important for optimizing performance across configurations, which can range from disaggregated data preprocessing~\cite{Audibert2023TFDataService,Zhang2025DistTrain,Anyscale2024Runway,Zhao2025OVERLORD} to multiprocessing or multithreading (with the removal of Python's GIL~\cite{Hira2025SPDL}) on training nodes. 
The choice of execution backend must not affect the training data sequence. %
\vldb{In this paper, we co-locate the data loader on the training node, as this is the standard industry setup that we observe in Datology's customer deployments.}

\revi{We present \zephon, \emph{to our knowledge, the first data loader to offer elastic determinism and efficient resumption for online, stateful $n$-to-$m$ foundation model data pipelines}.}
\zephon{} carefully manages the pipeline state, rather than representing it as a sample index.
Our contributions are as follows:

\begin{enumerate}[leftmargin=*,nosep]
\item \zephon{} introduces \emph{accumulators} which serialize stateful ordering decisions in a single thread while allowing stateless transformations to run in parallel on flexible execution backends called \emph{runners}. 
\revMRf{It generalizes \textsc{Mosaic Streaming}'s elastic determinism}  to online, stateful $n$-to-$m$ pipelines using a \emph{lane} abstraction, a logical sub-stream that produces a slice of the training data regardless of the underlying topology.

\item \zephon{} checkpoints stateful data loading pipelines and restores them with bounded cost. 
For this, it performs \emph{per-offset contributor tracking},  a mechanism that handles operators which reorder, merge, and spawn samples.

\item We release \zephon{} as an open-source foundation model data loading system, with a composable builder-pattern interface.
It ships with a toolbox of common foundation model operators.
\end{enumerate}

\section{Stateful \texorpdfstring{$n$}{n}-to-\texorpdfstring{$m$}{m} data loading pipelines}\label{sec:background}

\begin{figure}
    \centering
    \includegraphics[width=\linewidth]{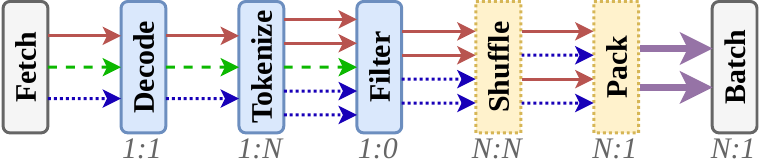}
    \caption{A non-indexable VLM preprocessing pipeline. Operators vary in cardinality (below the operator) and statefulness (blue solid/yellow dotted). Arrow color/pattern indicates the source, thick purple arrows are packed samples. } 
    \label{fig:back-vlm-pipeline}
    \vspace{-0.5cm}
    \Description{TODO!}
\end{figure}

Consider a VLM training corpus of image-caption pairs, interleaved documents, and text-only samples, as used for the MM1 family of models~\cite{McKinzie2024MM1}.
\Cref{fig:back-vlm-pipeline} shows a representative preprocessing pipeline (e.g., see the Qwen-VL model family~\cite{Wang2025OpenQwen2VL,Bai2025Qwen25VL,Bai2025Qwen3VL} and Eagle~2 model~\cite{Li2025Eagle2}).
The operators read records, decode text and images, plus, if necessary, apply vision transforms, apply the chat template and tokenize the sequence, shuffle, pack, and construct batches.

The operators in this pipeline differ in \emph{cardinality}.
Decoding and augmentation are $1$-to-$1$: one input record yields one output.
Tokenization with splitting is $1$-to-$n$: a long document becomes multiple subsequences.
Packing is $n$-to-$1$: several short sequences merge into one training example.
Filtering is $1$-to-$0$: records that fail a quality check are dropped.

Orthogonal to cardinality, many operators are \emph{stateful}.
Packing maintains partially filled bins across inputs.
Shuffle operators hold a buffer of pending samples.
Mixture enforcement tracks per-component counters to preserve dataset ratios.
Because of this state, the next output depends not only on the current input but on the history of prior inputs.

\revi{We call a pipeline \emph{indexable} if the source records contributing to its output at training step~$t$ can be located from~$t$ without evaluating the preceding input prefix.}
$n$-to-$m$ cardinality alone already breaks indexability.
When tokenization splits documents into a data-dependent number of subsequences, indexability is lost: the mapping from training step to source record depends on the length of all prior documents.
Statefulness compounds the \revi{problem, since resuming} from an arbitrary input offset requires reconstructing operator state.
Any change to the execution that alters how state accumulates (worker count, backend, cluster topology) silently changes the output sequence.

\section{Requirements overview}\label{sec:back-requirements}
  
A data loader for non-indexable pipelines must meet the same guarantees as for indexed pipelines: high throughput, composability, deterministic output, and efficient recovery.
However, the mechanisms indexed pipelines rely on (sample offsets, static partitioning, index-based resume) no longer apply.
We discuss why stateful, non-indexable pipelines make these requirements harder to satisfy:

\textbf{R1: High throughput and backend portability.}
Data loading must not bottleneck training~\cite{Mohan2021DataStalls}. %
In an indexable pipeline, parallel processing is straightforward using partitioning.
In a non-indexable pipeline, output boundaries are data-dependent, so the system must decouple parallel execution from the logical output order.
A data loader must keep throughput high and support different execution substrates (e.g., processes/threads) to achieve this throughput across diverse training workloads.%

\textbf{R2: Composability.}
Pipelines differ across teams and modalities.
In order to accommodate this, users should be able to compose operators freely and plug in custom transformations, rather than having a fixed, hardcoded pipeline for a single modality.
A data loader should be composable and should remain decoupled from any particular file format, storage backend, or training framework.

\textbf{\revMRf{R3: Elastic determinism across execution configurations.}}
\revi{Determinism in data loading supports debugging and controlled ablations, removes one confounder when diagnosing silent data corruption~\cite{Ma2025SDC,Gemini2025Multimodal}, and increasingly matters for compliance~\cite{EU2024AIActArt11}.}
\revMRf{If we view the loader as a stateful dataflow graph, its logical result is the ordered sequence of global batches consumed by training.
Because training is order-sensitive, elastic determinism is a correctness requirement: for a fixed logical pipeline, the result must remain identical when execution choices such as operator parallelism or execution substrate change.
It must also remain identical as the data-parallel degree changes, at constant global batch size.}
A data loader must provide elastic determinism even for non-indexable pipelines, where no static sample index exists to remap across topologies.

\textbf{R4: Efficient resume from checkpoint.}
The data loader state must be checkpointed with the model.
For an indexable pipeline, a sample offset suffices.
In a non-indexable pipeline, progress is entangled across operator state (bins, buffers, counters) with no compact representation.
Naively restoring such state by replaying the pipeline from the start causes recovery costs to grow with training progress.
A data loader must therefore support bounded-cost checkpoint recovery regardless of pipeline progress.

\section{\zephon's architecture and abstractions}
\label{sec:design}

\begin{figure*}
    \centering
    \includegraphics[width=\textwidth]{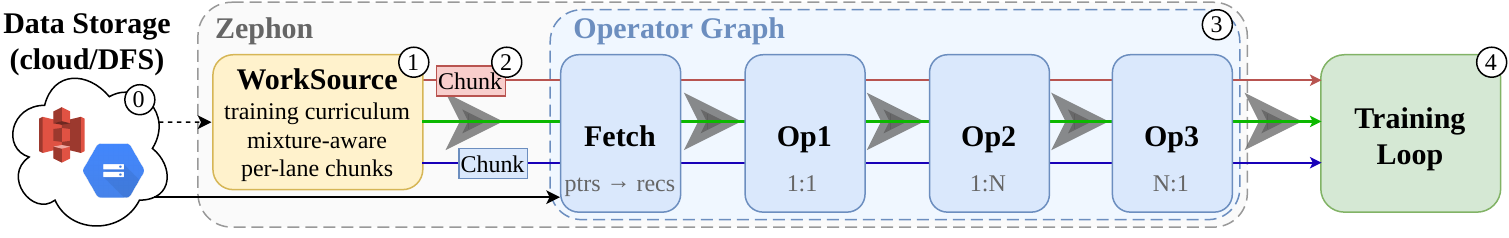}
    \caption{\zephon's high-level architecture. The WorkSource produces chunks for each lane (colored lines), which are converted into samples by the Fetch operator. Each lane holds an independent chunk stream and independent operator states.}
    \label{fig:design-arch}
    \Description{TODO!}
    \vspace{-0.5cm} 
\end{figure*}

We build \zephon, a data loader for training foundation models that addresses these requirements.
\zephon{} separates \emph{what} to train on from \emph{how} to process it: a \texttt{WorkSource} defines the \emph{training curriculum} (which datasets to use, in what proportions, in what order), and a pipeline of composable operators (\textbf{R2}) turns this into training batches.

\subsection{User interface}
\label{sec:execution:construction}

\revMRh{\zephon{} is available as a Python package and supports user-defined functions and arbitrary data payloads.}
\revi{We open-source the code,\footnote{\url{https://github.com/datologyai/zephon}} comprising ca.~20\,k LOC, excluding tests.}
Pipelines are defined via a builder-pattern API:

\begin{minted}[
frame=lines,
framesep=2mm,
baselinestretch=0.8,
bgcolor=LightGray,
fontsize=\footnotesize,
linenos
]
{python}
ws = StaticMixtureWorkSource(
    [Dataset.from_path("demo", "s3://example-dataset")], 
    mixture=MixtureSpec({"demo": 1.0}), chunk_size=1024)
pipeline = (Pipeline(ws)
    .tokenize(tokenizer_id="gpt2", parallelism=4)
    .pack_sequences(target_length=2048)
    .ensure_mixture()
    .batch(microbatch_size=8))
for batch in pipeline:
    train_step(batch.to_training())
\end{minted}

Lines 1--3 define the training curriculum via a \texttt{WorkSource}.
Lines 4--8 attach a chain of operators that turn raw data into training batches.
We now describe each of these abstractions.

\subsection{Component overview}
\label{sec:design:flow}

The overall data flow is shown in~\Cref{fig:design-arch}.
\zephon's mechanism for elastic determinism (\textbf{R3}) is the \emph{lane}.
The global data stream is partitioned into a fixed number of lanes configured by the user.
Each lane is a logical sub-stream that produces a deterministic, non-overlapping slice of the training data.
Each lane produces an independent, deterministic data stream with its own operator state.
\revMRf{Lanes generalize \textsc{Mosaic Streaming}'s canonical nodes~\cite{MosaicML2022Streaming} to non-indexable, stateful pipelines.}

\textbf{The \texttt{WorkSource}.} 
The \texttt{WorkSource}~\Circled{1} defines the training curriculum, e.g., by indexing the data~\Circled{0}.
For each lane, it produces fixed-size \emph{chunks} of $C$ sample pointers, ordered as $\langle p_0,\ldots,p_{C-1}\rangle$~\Circled{2}.
It does \emph{not} read the sample payloads.

\textbf{Fetch: from pointers to data.}
\zephon{} processes samples in a DAG of operators.
\revMRh{Every pipeline begins with the \emph{fetch} operator.}
For each pointer in the current chunk, it fetches the actual payload from the corresponding file and generates a record carrying it.
The record also carries metadata that uniquely identifies it within its lane, used for checkpoint resumption (\textbf{R4}/\Cref{sec:checkpoint}).

\textbf{Further processing.}
After fetch, records flow through the user-defined operator graph~\Circled{3}. 
Operators compose freely (\textbf{R2}), and the chain can contain arbitrary cardinalities, i.e., records can get filtered, split, etc.
After the final operator, \zephon{} yields the sample to the training loop~\Circled{4}.

\vldb{\textbf{Operators, not workers.} \zephon{} treats a data loading pipeline as a graph of operators that scale independently, rather than as workers that each run the whole pipeline  (as, e.g., in PyTorch's \texttt{DataLoader}).
This matters for two reasons. First, operators have different resource profiles.
For example, data fetching is IO-bound, while tokenization is CPU-bound.
Per-operator scaling avoids a single performance knob (\textbf{R1}).
Second, the degree of parallelism can change the output order in worker-based loaders, coupling a performance choice to training semantics.}

\subsection{Data model}
\label{sec:design:data}
\zephon{} follows the shard-based approach of  data loaders such as \textsc{Mosaic Streaming}~\cite{MosaicML2022Streaming} or \textsc{LitData}~\cite{Chaton2023LitData}.
A \emph{dataset} $\mathcal{D}$ is a named collection of data at a storage path, e.g., a local directory, an S3 bucket, or a distributed file system prefix.
A dataset is split into $k \geq 1$ \emph{shards} $\mathcal{D} = \{S_1, \ldots, S_k\}$, where each shard is a file containing a contiguous slice of the data.
A \emph{sample} is the smallest addressable unit within a shard: the $j$-th sample in shard $S_i$ of dataset $\mathcal{D}$ is addressed by the \emph{sample pointer} $p = (\mathcal{D}, i, j)$.
\zephon{} is not tied to a single file format; above the format-specific reader layer, the abstraction is always the same (\textbf{R2}): a dataset has shards, and each shard has a known number of samples.

\subsection{\texttt{WorkSource}}
\label{sec:design:worksource}

The \texttt{WorkSource} is \zephon's abstraction for decoupling the training curriculum from preprocessing.
This enables a model engineering team to iterate on downstream operators without changing the data composition, and vice versa to adjust training data without touching preprocessing code.

\textbf{The \texttt{WorkSource} protocol.}
A \texttt{WorkSource} is an object that produces a deterministic stream of chunks for each lane.
\zephon's current implementation specifies curricula at the file system level: mixtures reference datasets by storage path.
However, the abstraction is not tied to this.
Any backend that can translate a curriculum specification into chunks of sample pointers satisfies the protocol.
For example, one could implement a \texttt{WorkSource} backed by a centralized sample management server such as \mixtera~\cite{Bother2026Mixtera}, where curricula are expressed as SQL-like queries rather than file paths.

\textbf{Chunk structure.}
A chunk is a fixed-size sequence of sample pointers drawn from one or more mixture components (domains).
Given components $d_1, \ldots, d_m$ with target weights $w_1, \ldots, w_m$, each chunk contains pointers from each component proportional to its weight.
Iteration order within a chunk is controlled at two levels: \emph{inter-component}, where pointers are interleaved across components in a deterministic weighted round-robin, and \emph{intra-component}, where the \texttt{WorkSource} controls the order in which samples are yielded.
\revMRd{For intra-component ordering, the \texttt{WorkSource} can shuffle samples within and across shards before forming and assigning chunks to lanes.}
Because the curriculum is defined at chunk granularity, this approach naturally supports dynamic curricula~\cite{Chen2023Skillit,Bother2026Mixtera,Jiang2024ADO}: a \texttt{WorkSource} can adjust weights in between chunks.

\section{Execution model}
\label{sec:execution}

\begin{figure*}
    \centering
    \includegraphics[width=\textwidth]{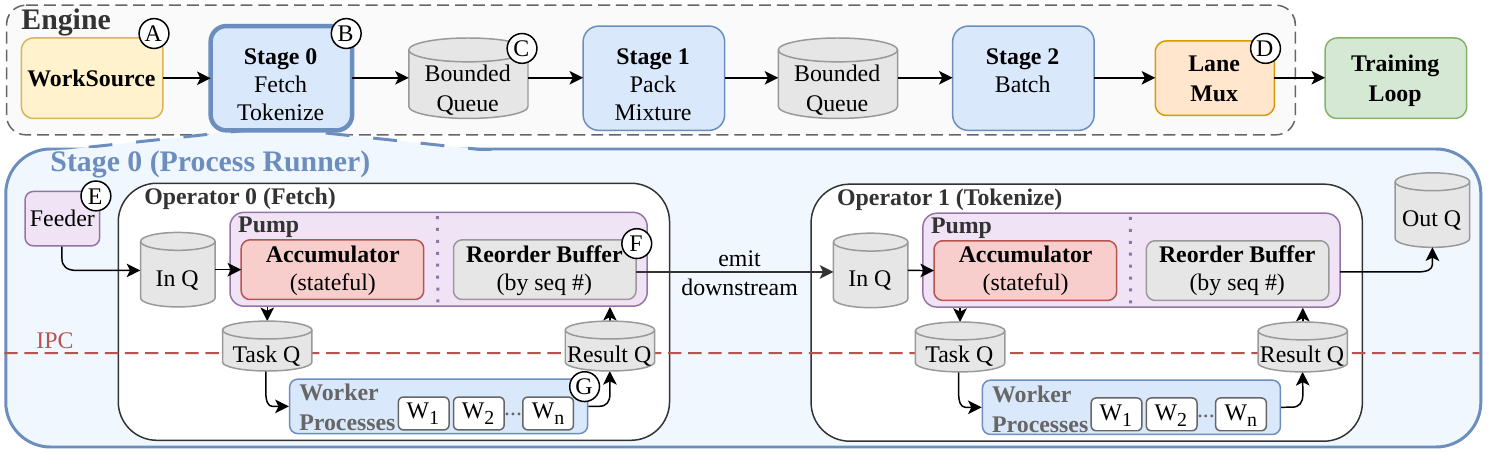}
    \caption{\zephon's runtime architecture. The engine feeds chunks into a pipeline compiled into stages, each assigned a runner. Within each stage, a single-threaded accumulator dispatches microbatches to parallel stateless workers.}
    \label{fig:execution-arch}
    \Description{TODO!}
    \vspace{-0.3cm} %
\end{figure*}

\Cref{fig:execution-arch} shows how \zephon{} executes pipelines to achieve high throughput (\textbf{R1}).
When training starts, \zephon{} compiles the operator graph into a \emph{plan} that groups operators into \emph{stages}, each assigned a \emph{runner}.
We describe the engine that orchestrates the data flow,  the accumulator model that enables parallel execution of stateful operators within each stage, and the runners that implement worker backends.

\subsection{Engine and pipeline execution}
\label{sec:execution:engine}

The \emph{engine} drives the compiled pipeline.
\vldb{\emph{Between} stages, execution is \emph{pull-driven}: stages are chained as Python iterators, and the training loop at the tail drives execution by requesting the next batch.
\emph{Within} a stage, execution is \emph{push-driven}: a dedicated pump per operator (\Cref{sec:execution:pump}) actively pushes records through its accumulator and workers.}
Prefetch buffers at multiple levels overlap computation with data movement.

\textbf{From \texttt{WorkSource} to stages.}
For each lane, the engine requests work chunks from the \texttt{WorkSource}~\Circled{A}.
The engine iterates over the pointer stream of each chunk and provides them to the first stage~\Circled{B}.
When a chunk is exhausted, the next chunk is requested.

\textbf{Inter-stage communication.}
Stages are connected by configurable bounded queues~\Circled{C}.
When a downstream stage is slow, its input queue fills up, and the upstream stage blocks on insertion.
This provides backpressure that prevents unbounded memory growth.

\textbf{Consumer handoff.}
Data from all lanes (\Cref{sec:design:flow}) flow through the system.
At the tail of the pipeline, the \emph{lane multiplexer}~\Circled{D} collects fully processed data from all lanes.
It buffers per-lane output and emits data in deterministic round-robin order across lanes.
Lanes may produce at different rates (e.g., if one lane's samples are more expensive to process).
The lane multiplexer buffers faster lanes while waiting for the next lane in the round-robin.
This is key to elastic determinism (\Cref{sec:determinism}).
The training loop consumes batches through a standard Python iterator, with a prefetch buffer to overlap data loading with GPU computation.

\subsection{Stage execution and the operator pump}
\label{sec:execution:pump}

Operators like image decoding are expensive and must run in parallel, without blocking the  training process.
\zephon{} supports this, even for stateful operators, by splitting each operator into a stateful \emph{accumulator} and stateless \emph{workers}, orchestrated by a \emph{pump}.

\textbf{Stage structure.}
A stage with $k$ operators spawns one \emph{feeder}~\Circled{E} and $k$ pumps, connected by bounded queues.
The feeder runs in a dedicated thread, pulling records from upstream and pushing them into the first operator's input queue.
Each operator has a dedicated pump~\Circled{F} running in its own thread.
The pump processes records through the operator-specific accumulator and parallel workers~\Circled{G}, then pushes results into the next operator's input queue, or, for the last operator, into the \emph{stage output queue}.
Each stage is therefore simultaneously consumer and producer: it pulls from upstream via the feeder and exposes an iterator downstream, while internally all forwarding is push-driven.
All queues are bounded, so backpressure propagates end-to-end. 
Between operators, a slow consumer fills its input queue, blocking the upstream pump; between a pump and its workers, bounded in-flight queues stall the pump.

\textbf{The pump loop.}
Each pump runs the loop in~\Cref{alg:pump}.
It pulls microbatches from its input queue and feeds them to the operator-specific accumulator.
A \emph{microbatch} is a small group of records processed and forwarded as a single unit, amortizing per-record overhead (queue operations, serialization, cross-process transfer).
The accumulator runs within the pump and is stateful: it receives incoming microbatches and reorganizes them as needed for the operator.
For example, a packing accumulator holds partially filled bins until they reach the target length, and emits a microbatch of samples that should be packed together.
The accumulator may not mix records across lanes.

When the accumulator emits a microbatch, the pump dispatches it to a pool of workers that apply the operator's transformation.
Workers may complete out of order; a \emph{reorder buffer} reassembles results using the sequence number assigned at dispatch. %
\textsc{Dispatch} and \textsc{DrainResults} are operations specialized by the runner.

\begin{algorithm}[t]
\small
\caption{Pump loop for a single operator.}
\label{alg:pump}
\KwIn{operator state $s$, worker backend, output queue $q_\text{out}$}
\While{$\mathit{mb} \gets s.\text{input\_queue.get()} \neq \textsc{Stop}$}{
    \textsc{DrainResults}($s$, $q_\text{out}$)\label{line:drain}\tcp*{reorder+forward}
    $\mathit{ready} \gets s.\text{accumulator.push}(\mathit{mb})$\;
    \ForEach{microbatch $b \in \mathit{ready}$}{
        $s.\text{seq} \gets s.\text{seq} + 1$\;
        \textsc{Dispatch}($b$, $s.\text{seq}$, worker backend)\label{line:dispatch}\;
        \textsc{DrainResults}($s$, $q_\text{out}$)\tcp*{keep workers fed}
    }
}
$s.\text{accumulator.flush}() \to$ dispatch remaining\;
drain all in-flight results into $q_\text{out}$\;
$q_\text{out}.\text{put}(\textsc{Stop})$\;
\end{algorithm}

\subsection{The runner}\label{subsec:exec-runner}

Each stage is assigned a runner that controls the execution of operators. 
\revMRh{A runner determines how worker functions are executed and thus provides the interchangeable execution substrate (\textbf{R1}).}
Currently, \zephon{} supports a process and a thread runner \vldb{that co-locate data loading on the same node as the training process.}
\vldb{\emph{Disaggregated data preprocessing} that moves preprocessing operations to separate CPU nodes is mostly used in recommendation systems~\cite{Zhao2022MetaRM} and proprietary generative video models~\cite{Anyscale2024Runway}.
Mainstream training frameworks such as \textsc{Torchtitan} and \textsc{Megatron} all co-locate the data loader.
Even systems built for disaggregation, such as \textsc{Pecan}, co-locate by default and offload only when profiling identifies a local-CPU bottleneck~\cite{graur2024pecan}.
We leave a detailed treatment of the challenges of providing elastic determinism under disaggregation to future work.
}

\subsubsection{The process runner}
\label{sec:execution:process-runner}

Python's Global Interpreter Lock (GIL) blocks threads from running CPU-bound code in parallel.
\zephon's process runner therefore executes workers in separate processes. 
\textsc{Dispatch} (line~\ref{line:dispatch} of \Cref{alg:pump}) serializes the microbatch via \texttt{pickle} and sends it over a cross-process queue.
\textsc{DrainResults} (line~\ref{line:drain}) deserializes results from a corresponding return queue.

\textbf{Operators and IPC.} Process-based execution in an operator model is more challenging than in a monolithic-worker model as used by, e.g., PyTorch's \texttt{DataLoader}.
At startup, each operator's stateless transformation must be serialized and sent to its worker processes.
At runtime, every microbatch crosses a process boundary on each dispatch and return, unlike monolithic-worker systems, where only the final output leaves the process.
Operator-based loaders like \textsc{Cedar}~\cite{Zhao2024cedar} address this by fusing adjacent operators to reduce cross-process transfers.
\zephon{} minimizes transfer cost through shared-memory coalescing. %

\textbf{IPC and shared-memory coalescing.}
\revi{The process runner's queues use OS pipes, where serializing large payloads is costly and can saturate pipe buffers.
PyTorch reduces this cost for tensors by placing each tensor in a POSIX shared-memory (shm) segment.
However, per-tensor allocations and system calls remain expensive, many tensors can exhaust the per-process file-descriptor (FD) limit, and non-PyTorch payloads still traverse the pipe.
\zephon{} addresses these limitations by coalescing all large payloads in a microbatch before enqueueing.
It groups tensors by \texttt{dtype}, allocates one contiguous shm region per dtype, and copies each tensor into its slice; large byte buffers and NumPy arrays are packed by the same mechanism.
For these payloads, the pipe then carries only lightweight metadata (shapes, offsets, and shm handles), allowing the receiver to reconstruct them as zero-copy views.
Coalescing thus reduces the number of segments from one per payload to one per dtype per microbatch, cutting allocation overhead and FD pressure while extending shared-memory transport beyond PyTorch tensors.}

\subsubsection{The thread runner}
\label{sec:execution:thread-runner}

Thread-based execution avoids the serialization and shared-memory management complexity of the process runner entirely.
Threads share the same address space and achieve true parallelism for operators implemented as C extensions or that otherwise release the GIL, as pioneered by \textsc{Spdl}~\cite{Hira2025SPDL}.
Alternatively, free-threaded Python (3.13t/3.14t) removes the GIL, making threads viable for all operators.

The thread runner uses a thread pool within the current process.
\textsc{Dispatch} (line~\ref{line:dispatch}) submits the microbatch to the thread pool; \textsc{DrainResults} (line~\ref{line:drain}) reads results from an in-process queue.

For each operator in the stage, the thread runner keeps a pool of pre-initialized operator instances.
A worker thread checks out an instance for the duration of a microbatch.
Python's \texttt{Future} callbacks run synchronously on the caller when the future is already complete; since the pump is the sole consumer of the result queue, such a callback writing to a full queue would deadlock.
\zephon{} avoids this by stashing synchronous results in a thread-local buffer that the pump drains inline after each dispatch.

\subsection{\vldb{User obligations and system guarantees}}\label{sec:execution:contract}

\vldb{In most cases, the execution machinery above is almost entirely hidden from users.
\zephon{} comes with a toolbox of operators and \texttt{WorkSource} implementations that are deterministic by construction, and users typically just compose them (\Cref{sec:execution:construction}).
Users may also insert stateless \texttt{map} or \texttt{filter} UDFs, which \zephon{} wraps into operators automatically.
Only advanced users who implement custom
stateful $n$-to-$m$ operators must reason about determinism (\Cref{sec:determinism}) and the sample lifecycle (\Cref{sec:checkpoint}).}

\vldb{\textbf{State.} When writing a custom operator, users need to separate state from the stateless processing function.
State needs to be managed per lane.
\revMRd{For example, the shuffle-buffer operator provides additional shuffling for each lane, supplementing the \texttt{WorkSource}'s shuffle before lane assignment (\Cref{sec:design:worksource}).} 
\revMRg{Generally, accumulator decisions must depend only on the input sequence, not on how upstream operators group records into microbatches.}
}

\vldb{
\revMRh{
In addition, operators must seed randomness from the contents of each microbatch rather than use a global RNG.
For example, \zephon's shuffle-buffer implementation does this, ensuring identical output regardless of which worker processes the microbatch.}
}

\vldb{\textbf{Auto-validation.} As these contracts cannot be enforced by Python's type system, \zephon{} has a validation harness that automatically probes each user-defined operator, both statically on its AST and by running it on synthetic samples.
\zephon{} flags operators whose output is non-deterministic or that wrongly handle sample metadata to help users catch these bugs.}

\revMRf{\textbf{Trainer consumption.} For a fixed global batch size of $N$ records and $L$ lanes, consider an execution with \revMRe{data-parallel degree $G$}, per-replica training microbatch size $m$, and $a=N/(Gm)$ gradient-accumulation steps.
\revMRe{With TP, PP, or EP, lanes map to DP coordinates rather than global ranks; the training framework decides which ranks materialize, replicate, or shard the corresponding batches.}
\zephon's elastic determinism requires $G\mid L$ and $L/G\mid a$; the latter is equivalent to $L\mid Ga$, where $Ga=N/m$.
The first condition assigns equally many lanes to every data-parallel replica; the second ensures that every lane contributes a whole number of microbatches to every global batch.
\revMRe{Users choose $L$ as a common multiple of the anticipated data-parallel degrees while ensuring that each intended $m$ divides $N/L$.
The fixed $L$ caps the data-parallel degree at $L$; larger values support more scale points but increase per-lane bookkeeping.}
Other configurations with $G\le L$ still execute deterministically under a fixed topology, but \zephon{} does not guarantee identical global batches across topology changes.
}

\revMRf{Subject to these obligations, \zephon{} guarantees an identical global batch sequence across supported execution configurations.}

\section{Deterministic and elastic execution}
\label{sec:determinism}

\revMRf{This section formalizes elastic determinism as a correctness guarantee over \zephon's stateful dataflow graph and shows that its ordered sequence of global batches is invariant across supported physical executions (\textbf{R3}).}

\subsection{Determinism by induction}
\label{sec:determinism:det}

We show by induction that the output of a \zephon{} pipeline is deterministic.
Intuitively, the accumulator is the sole component that decides ordering; all others mostly forward, so the overall stream is deterministic.

\textbf{Base case: single-operator determinism.}
Consider a single operator executed by a runner (\Cref{subsec:exec-runner}).
\revMRg{It follows the rules in \Cref{sec:execution:contract}.}
The pump (\Cref{alg:pump}) receives records in a fixed order, makes stateful decisions (grouping, buffering, emission) in the accumulator, and then tags each outgoing microbatch with a monotonic sequence number.
Workers execute only their stateless processing function and may complete out of order.
The reorder buffer then reassembles results by sequence number before forwarding them downstream.

\textbf{Inductive step: pipeline determinism.}
Let $O_1, \ldots, O_k$ be operators with deterministic output.
Then $O_{k+1}$ receives deterministic input and, analogous to the base case, produces deterministic output.
Stages and operators within a stage are connected by order-preserving FIFO queues.
The full pipeline is therefore deterministic by induction.

\subsection{Elastic determinism}
\label{sec:determinism:elastic}

\revMRf{Let $\Pi=(\mathbf I,\mathcal P,N)$ denote a fixed semantic pipeline specification.
Here, $\mathbf I=\langle I_0,\ldots,I_{L-1}\rangle$ contains the canonical ordered input stream of every lane, including flush sentinels (\Cref{sec:checkpoint:nonmonotone}), $\mathcal P$ is the operator graph, and $N$ is the global batch size.
This definition is semantic: an implementation or configuration change is comparable only if it preserves $\mathbf I$ and $\mathcal P$.

The trainer consumption contract in \Cref{sec:execution:contract} ensures that $N/L=ma/(L/G)$ is an integer; let $q=N/L$.
Let $\mathcal E(\Pi)$ denote the supported physical executions of $\Pi$.
Executions in $\mathcal E(\Pi)$ may differ in runner, operator parallelism, data-parallel degree, and lane-to-replica assignment.
Training microbatch size may also differ as it only regroups the same delivered records.
An execution $e$ with data-parallel degree $G_e$ and training microbatch size $m_e$ belongs to $\mathcal E(\Pi)$ only if $G_e\mid L$ and $m_e\mid q$.

For $e\in\mathcal E(\Pi)$, let $S_\ell(\Pi,e)=\langle s_{\ell,0},s_{\ell,1},\ldots\rangle$ denote the ordered record sequence delivered for lane~$\ell$, flattening the output if a \texttt{Batch} operator is present.
The contribution of lane~$\ell$ to optimizer step~$t$ is $b_{\ell,t}(\Pi,e) = \langle s_{\ell,tq}, \ldots, s_{\ell,(t+1)q-1} \rangle$. 
The corresponding logical global batch is $B_t(\Pi,e) = \langle b_{0,t}(\Pi,e), \ldots, b_{L-1,t}(\Pi,e) \rangle$.
\zephon{} provides \emph{elastic determinism} iff f.a. $ e,e'\in\mathcal E(\Pi)$ and f.a. $t: B_t(\Pi,e)=B_t(\Pi,e')$.
We omit $\Pi$ and $e$ below.
}

\revMRf{\textbf{Lanes as the unit of elasticity.}
Each lane's flattened output sequence is independent of physical execution parameters such as runner or operator parallelism.
Lanes are therefore the topology-independent units of the output.
With data-parallel degree $G$, each replica owns $L/G$ lanes (with $G\mid L$).
Changing $G$ reassigns lanes to ranks but does not change their input streams, operator state, or ordered outputs. 
It remains to show that the per-rank outputs are consumed such that every optimizer step receives the same contribution $b_{\ell,t}$ from every lane.}

\revMRf{\textbf{Round-robin execution and output multiplexing.}
Samples from all lanes owned by a rank flow through the same pipeline.
At the source, the engine feeds the lanes in round-robin order.}
However, as lanes can produce output at different speeds, the raw pipeline output arrives in unpredictable lane order.
\revMRf{
The \emph{lane multiplexer} (\Cref{sec:execution:engine}) restores the source ordering by buffering outputs in per-lane queues and emitting from the owned lanes in strict round-robin order.
Faster lanes wait until their turn.

Let $m$ be the number of records in a training microbatch\footnote{Note that \zephon's internal microbatching does not need to align with the training microbatches emitted by the batch operator.}.
For a supported execution, $m\mid q$.
Before each optimizer step, every rank consumes $N/(Gm)$ training microbatches from \zephon{} and accumulates their gradients.
Since a rank owns $L/G$ lanes, this corresponds to $\frac{N/(Gm)}{L/G} = \frac{N}{Lm} = \frac{q}{m}$ complete multiplexer rounds. 
Each owned lane therefore contributes $q/m$ training microbatches, containing exactly the $q$ records in $b_{\ell,t}$, to optimizer step~$t$.
Thus every optimizer step advances every lane by exactly $q$ records.
The calculation depends only on the number of lanes owned by each rank, not on which lanes it owns, so the lane-to-rank assignment does not affect $B_t$.}

\textbf{Lane reassignment preserves global batches.}
\revMRf{The lane multiplexer and the trainer's accumulation schedule ensure that each global batch contains exactly one contribution $b_{\ell,t}$ from every lane.
Because each lane's stream is deterministic and lane-local, $b_{\ell,t}$ is the same regardless of which replica owns lane~$\ell$.
Changing $G$ therefore changes only where each contribution is produced, not $B_t$; together with \Cref{sec:determinism:det} and deterministic replay in
\Cref{sec:checkpoint}, this establishes elastic determinism across all executions in $\mathcal E(\Pi)$.

Note that \zephon's guarantees end at $B_t$: different GPU topologies may change the training framework's float accumulation and reduction order, so identical global batches
do not imply bit-identical losses or model states.
\zephon{} may execute configurations outside $\mathcal E(\Pi)$ (e.g., $G\nmid L$ or $m\nmid q$) deterministically under a fixed topology.
}

\textbf{Example.}
\revMRf{With $L=4$ lanes and $q=m$, two GPUs own lanes $\{0,1\}$ and $\{2,3\}$ and emit two microbatches per step, round-robin over their lanes. With four GPUs, each owns one lane and emits one. Either way, $B_t = \langle b_{0,t},b_{1,t},b_{2,t},b_{3,t}\rangle$.}

\section{Checkpointing and resumption}
\label{sec:checkpoint}

\zephon{} checkpoints and resumes the pipeline without replaying from the start (\textbf{R4}).
Our key insight is that only \emph{in-flight chunks}, i.e., chunks whose samples have entered the pipeline but have not all been delivered to the training loop, need to be saved alongside the \texttt{WorkSource} state and replayed on resumption.
\revMRb{A checkpoint stores their sample pointers, not payloads, so}
checkpoint size is bounded by the number of in-flight chunks\revMRb{, which backpressure caps regardless of training progress (\Cref{sec:execution:engine}).}
The important question is when a chunk can be \emph{evicted}, i.e., how the engine knows that all its samples have been delivered.

\subsection{Cursors and contributors}
\label{sec:checkpoint:identity}

A record's \emph{cursor} uniquely identifies it.
\vldb{Orthogonally, \emph{contributor references} track which source records a record contains.}

\textbf{Cursors.}
\revMRh{Each record has a \emph{cursor} $(\mathsf{chunk},\, \mathsf{offset},\, \mathsf{lineage})$ that uniquely identifies it within its lane.}
\vldb{The first two fields are the \emph{base offset} of the cursor, i.e., the source chunk position the cursor is derived from.
$\mathsf{lineage}$ is a path of integer indices that grows as fan-out operators derive new records, distinguishing records that share a base offset.
For example, splitting a document into three sequences yields lineages $(0,)$, $(1,)$, $(2,)$; further splitting child~1 produces $(1,0)$, $(1,1)$, etc.
Fan-in operators like packing choose one \emph{primary} cursor for the output, derived from one source record.
This primary cursor serves as the packed record's replay identity, while all source records, including the primary one, are tracked for eviction through contributor references (below).}

\revMRg{A checkpoint, in addition to the in-flight chunks and \texttt{WorkSource} state, stores one \emph{replay cursor} per lane.
For a batched pipeline, the replay cursor is batch-aligned: it is the cursor of the last record, in pipeline order, in the last batch delivered from that lane.
For an unbatched pipeline, it is the cursor of the last delivered record.}

\textbf{Contributors.}
\vldb{A record has one cursor as its replay identity, but a fan-in operator like packing may assemble its payload from \emph{multiple} source records.
Each record therefore carries one \emph{contributor reference} per source record, each storing that source record's cursor and an \texttt{is\_last\_child} flag.
A $1$-to-$1$ record has one contributor: its own cursor.
The base offset $(\mathsf{chunk},\, \mathsf{offset})$ of a cursor is the granularity at which we track completion.
A base offset can be closed when a delivered contributor for it carries \texttt{is\_last\_child}.
A single packed record can close several base offsets at once.
}

\subsection{Replay}
\label{sec:checkpoint:replay}

On resumption, the engine restores from the checkpoint in two phases.
In Phase~1, it \emph{replays} the in-flight chunks through the pipeline.
Because the pipeline is deterministic, replay produces identical output, including an identical accumulator state.
A \emph{replay filter} operator, automatically inserted during pipeline compilation, drops the already delivered prefix.
It compares each record's cursor against the stored replay cursor and suppresses everything up to and including the match.
In Phase~2, the engine fetches new chunks from the restored \texttt{WorkSource} and normal execution continues.

\textbf{Elastic resumption.}
In distributed training, each rank's engine owns a subset of lanes.
A complete checkpoint must contain state for \emph{all} $L$ lanes so that any rank assignment can restore from it. \revMRb{Per-rank state grows sublinearly with the lanes a rank owns.}
Unlike data loaders where each node independently saves and restores its own state, \zephon{} must aggregate per-rank state into a global, self-contained checkpoint because on elastic resume, a rank may receive lanes it did not own previously and needs their state.
When resuming on a different number of GPUs, each rank loads the merged checkpoint, takes ownership of its assigned lanes (\Cref{sec:determinism:elastic}), and replays its in-flight chunks \revMRb{independently, without further coordination}.

\begin{figure}
  \includegraphics[width=0.92\linewidth]{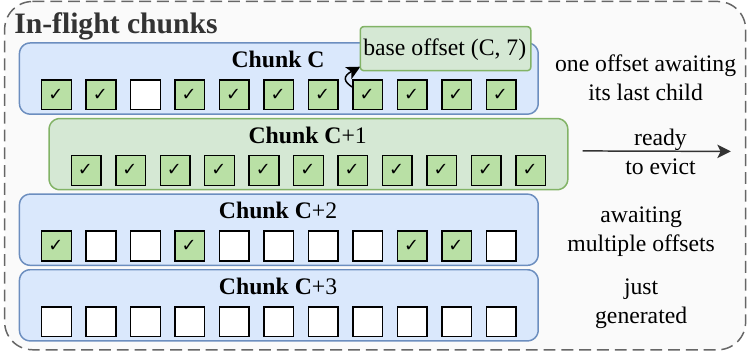}
    \caption{\vldb{Per-chunk completion bitmaps for in-flight chunks.
    Each cell is one base offset; it closes when its last-child contributor is delivered.
    Non-monotone operators reorder samples across chunks, so chunks close out of order, and a chunk is evicted only once its bitmap is full.}}
  \label{fig:bitmap}
  \vspace{-0.5cm}
\end{figure}

\subsection{Eviction}
\label{sec:checkpoint:eviction}

We now discuss when a chunk can be evicted.

\textbf{Monotone eviction.} We call a pipeline \emph{monotone} if every operator both preserves cursor order (within each lane, records are emitted in nondecreasing cursor order) and carries no history-dependent state beyond its current buffer.
In such a pipeline, eviction reduces to a simple watermark: if a record from chunk~$c$ reaches the training loop, all chunks before~$c$ are complete and can be evicted, because no operator's future output depends on them.

\label{sec:checkpoint:nonmonotone}

\textbf{Non-monotone eviction.}
Operators like packing reorder records across chunk boundaries, breaking the watermark invariant.
For example, a packed record may combine fragments from chunks~5 and~7, and chunk~5's last sample may be delivered \emph{after} a record from chunk~7.
Evicting chunk~5 based on seeing a sample from chunk~7 would lose data.

\vldb{Hence, in the general non-monotone case, the engine tracks per-offset progress.
It maintains a \emph{completion bitmap} (\Cref{fig:bitmap}), marking offset~$j$ of chunk~$c$ as closed when a contributor with base offset $(c, j)$ and the last child flag set is delivered to the training loop.}

A chunk can be evicted only once all its offsets are closed.
Completion bitmaps are not persisted in \revi{checkpoints; rather, they are rebuilt} during replay.
When a filter drops a record that is the last child for some base offset, the offset would never close.
Hence, operators emit a \emph{tombstone}, i.e., a non-payload sentinel carrying only a closing contributor reference.

The problem of non-monotone pipelines is deeper than  reordering.
A history-dependent accumulator state, e.g., the packing accumulator's bin layout, depends on the \emph{full history} of inputs, not just the records currently buffered.
After a chunk's samples have all left the bins, their influence on bin placement persists.
Crucially, two accumulator instances that start from different states may never converge, even on identical subsequent input.
For example, consider a bin packer with 2~bins of capacity~10 and an input stream of elements of size~3.
Starting from remaining-capacity states $(10, 10)$ vs.\ $(10, 5)$, both enter periodic emission cycles of the same period but permanently phase-shifted.

This means that completion bitmaps alone are not sufficient.
Replaying from a checkpoint would start the accumulator from an empty state instead of its true state during the checkpoint, producing different output and breaking deterministic replay.

\textbf{Flush sentinels and segments.}
\zephon{} solves this by injecting a \emph{flush sentinel} in the stream every $K$ chunks per lane in non-monotone pipelines.
When a flush sentinel reaches an accumulator, it emits all buffered data and additionally \emph{resets} internal state.

Conceptually, the flush sentinels divide the source stream into \emph{segments} of $K$ chunks each, where segment $S_i$ spans the chunks $[iK,\; (i{+}1)K)$.
Flush sentinel positions are stored in the checkpoint and re-injected at the same positions on replay. %
Within a segment, the accumulator builds up state normally.
At the segment boundary, the flush resets it to a clean state, making the next segment independent of all prior segments.
For monotone pipelines, no flushes are needed.
The resulting stream is deterministic for a fixed $K$. %

\vldb{\textbf{The segment-size tradeoff.}
\revMRb{A smaller $K$ allows chunks to be evicted more frequently.
However, each flush forces operators such as the packer to emit partially filled bins, which may increase padding waste.
Changing $K$ has little effect on checkpoint state because backpressure limits how far the pipeline reads ahead
(\Cref{subsec:eval-checkpointing}).}}

\textbf{Flush discipline.}
Non-monotone operators must flush immediately when the flush sentinel arrives.
Stalling would let post-boundary records mix with pre-boundary buffered state, creating an accumulator state that replay from empty cannot reconstruct.
Furthermore, a stalled flush sentinel delays the reset for downstream operators, causing their flushes to occur in divergent state.

The exception is the \texttt{Batch} operator, which stalls the flush sentinel, allowing post-boundary records to complete a partial batch.
This is safe because \texttt{Batch} preserves cursor order and has no history-dependent state beyond its buffer.
The replay filter is placed immediately before \texttt{Batch} during compilation; on resume, the consumed prefix is dropped before batching, so \texttt{Batch} rebuilds from an empty buffer at the correct batch-aligned cut.

\begin{lemma}[Segment eviction safety]
\label{lem:eviction}
Segment $S_i$ can be safely evicted when these two conditions hold:
\begin{enumerate}[nosep]
    \item All offsets in all chunks of $S_i$ are closed (per-chunk completion bitmaps full).
    \item The replay cursor does not reference a chunk in $S_i$.
\end{enumerate}
\end{lemma}
\begin{tcolorbox}[colback=gray!5, colframe=gray!30, boxrule=0.4pt, arc=1pt, left=4pt, right=4pt, top=2pt, bottom=2pt,before skip=4pt,after skip=4pt]
\begin{small}
\begin{proof}[Proof sketch]
The flush sentinel injected after $S_i$ drains and resets all accumulators, \underline{so no future output depends on $S_i$}.
Condition~1 implies any \texttt{Batch}-stalled flush sentinel has already released: offsets close only when last-child contributors are delivered in complete batches, and \texttt{Batch} preserves cursor order, so all $S_i$ records have been emitted before condition~1 holds.
\revMRg{The batch-aligned replay cursor (\Cref{sec:checkpoint:identity}) identifies exactly this consumed prefix.}
There is no incorrect early eviction due to stalling.
\revMRg{Since segments are evicted in order, condition~2 ensures the replay cursor is past $S_i$, so replay begins from the next segment's clean state.}
\end{proof}
\end{small}
\end{tcolorbox}

For non-monotone pipelines, eviction is atomic per segment as chunks within a segment share accumulator history.

\textbf{Example.}
Consider a non-monotone pipeline with $K = 2$.
Lane~0 has in-flight chunks 4--7 spanning segments $S_2 = \{4, 5\}$ and $S_3 = \{6, 7\}$, with a flush sentinel between them.
The replay cursor references chunk~6.
At checkpoint time, all offsets in chunks 4 and 5 are closed (condition~1) and the replay cursor does not reference a chunk in $S_2$ (condition~2), so the engine evicts $S_2$.
The checkpoint saves only \revMRb{the pointers of} chunks 6--7, the \texttt{WorkSource} position past chunk~7, the replay cursor, and the stored boundary position~6 (marking the start of $S_3$).

\section{Evaluation}

We evaluate \zephon{} to answer the following questions:

\begin{enumerate}[leftmargin=*,nosep]%
    \item How competitive is \zephon's throughput, how do its knobs affect scaling, \revMRc{and what do its guarantees cost}? (\textbf{R1})
    
    \item \revMRb{How do checkpoint size and recovery cost scale with training progress and lane count? (\textbf{R4})}

    \item \revi{How does elastic determinism affect training loss and downstream accuracy across GPU counts? (\textbf{R3})}
\end{enumerate}

\noindent\textbf{Setup.} All performance benchmarks use an AWS \texttt{r8id.48xlarge} instance (Intel Xeon 6975P @ 2.70\,GHz, 96 physical cores, 1.5\,TiB RAM, $3\times3.5$\,TB NVMe in RAID-0) running Ubuntu 24.04.
\revMRb{The instance is a single-tenant (dedicated) CPU host running the data loader and all its workers. 
The checkpointing experiments use eight parallel processes on this host. 
Training experiments use 8–64 H100 GPUs, with the data loader co-located with each training rank.}
To minimize variance, we disable SMT, C-state transitions, ASLR, and the NMI watchdog.

\subsection{Throughput}
We compare \zephon{} to other data loaders and evaluate how its tuning knobs affect performance.
Rather than end-to-end model training throughput, we use raw data loader throughput as the evaluation metric, as this is model-independent.
For a given training scenario, to avoid data stalls, the data loader must sustain a throughput higher than the model’s peak data consumption rate~\cite{Mohan2021DataStalls,Zhao2024cedar}.

For each scenario, we warm up the data loader until it reaches steady state, and then measure throughput over a fixed number of steps. 
We run every benchmark with three random seeds and report the mean.
We store all data in the \textsc{LitData} data format~\cite{Chaton2023LitData}, in shards of roughly 64\,MiB size.
We store all data on the local NVMe raid to benchmark the performance of the data loader instead of cloud storage throughput.
\vldb{Cloud storage throughput has high variance, and the intention of these benchmarks is to show the limits of the data loaders and their processing pipeline.
Fetching data from the cloud would mostly show bandwidth limitations, not the scalability of the implementation.}
We consider two benchmarks, \textit{fetch-and-batch} (\Cref{subsubsec:fetchnbatch}), which loads data like pretokenized strings from disk, and a \textit{VLM pipeline} (\Cref{subsubsec:vlm-pipeline}).

\subsubsection{Fetch-and-batch}\label{subsubsec:fetchnbatch}

    \begin{figure*}
    \begin{minipage}[t]{0.377\textwidth} %
      \centering
      \includeinkscape[width=\linewidth]{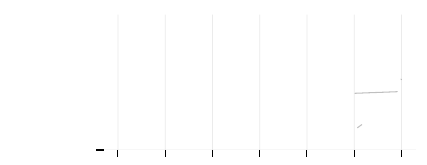}\\
      \includeinkscape[width=\linewidth]{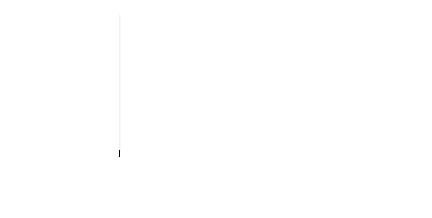}
      \subcaption{Comparison across loaders and parallelism}
      \label{fig:5-overview}
    \end{minipage}%
    \begin{minipage}[t]{0.300\textwidth} %
      \centering
      \includeinkscape[width=\linewidth]{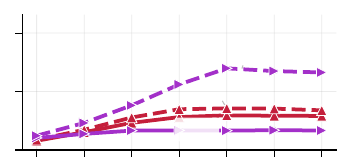}\\
      \includeinkscape[width=\linewidth]{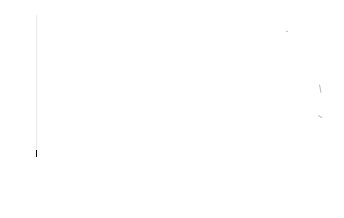}
      \subcaption{Effect of GIL removal}
      \label{fig:5-multithread}
    \end{minipage}%
    \begin{minipage}[t]{0.300\textwidth} %
      \centering
      \includeinkscape[width=\linewidth]{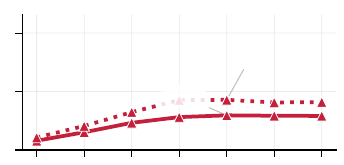}\\
      \includeinkscape[width=\linewidth]{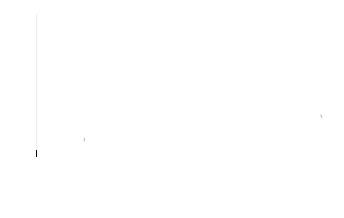}
      \subcaption{Impact of SHM coalescing}
      \label{fig:5-shm}
    \end{minipage}

    \caption{Fetch-and-batch throughput for text (top) and VLM (bottom) data with \zephon's process (\textsc{Z-Proc}) and thread (\textsc{Z-Thread}) runners. On text data, \zephon{} is sufficiently fast for stall-free training; on VLM data, it outperforms all baselines.}
    \label{fig:eval-tput}
    \vspace{-0.5cm}
  \end{figure*}

In this benchmark, we load raw sample payloads and create batches.
While this scenario involves minimal processing, it stress-tests the overhead of \zephon's operator machinery in a setting where workers have a structural advantage, yet highlights the flexibility of the operator-based architecture.

\textbf{Setup details.} Besides \zephon, we test the \textsc{LitData} data loader, a fork of \textsc{Mosaic Streaming} (short \textsc{Streaming}) with support for \textsc{LitData} shards, and \textsc{Google Grain} (short \textsc{Grain}). 
The text samples are sequences of 4\,096 tokens.
The VLM samples are \emph{multimodal documents}, i.e., PyTree-serialized \texttt{dict}s containing metadata, text (as strings), and a variable number of images (as packed \texttt{bytes}), from the MAmmoTH-VL dataset~\cite{Guo2025MAmmoTHVL}.
We shuffle the sample order \revMRd{in the \texttt{WorkSource}.}
For \zephon, we vary the \revi{degree of parallelism (DOP)} of the fetch operator, and for the other worker-based loaders, we vary the number of workers.
We report results for batch size~32.

\textbf{Overview.} The results are displayed in \Cref{fig:5-overview}.
For text data, the worker-based baselines perform very well.
\textsc{Grain} peaks at 201\,k\,samples/s at 16~workers, benefiting from two-stage prefetch both within the underlying dataset and its workers.  
\textsc{LitData} and \textsc{Streaming} reach 120\,k and 100\,k, respectively.
All three show diminishing returns or regression beyond their optimal worker count.
\zephon's process runner saturates at 59\,k\,samples/s (DOP${\geq}$8).
Its thread runner, constrained by the GIL, plateaus at just 34\,k\,samples/s despite avoiding IPC entirely.
For small text payloads, workers that combine fetching and batching in a single process have a structural advantage over \zephon's finer-grained operator model.

VLM data, where samples are variable-size dicts containing text and multiple images, favors \zephon.
\textsc{LitData} and \textsc{Streaming} fail to scale beyond
${\sim}$2\,k\,samples/s regardless of the number of workers;
\textsc{Grain} scales further to 5.8\,k at 64~workers.
\zephon's process runner reaches 16\,k\,samples/s, a
$2.7\times$ improvement over the best baseline, and its
(GIL-limited) thread runner reaches 10\,k.

\revMRc{\textbf{Cost of the guarantees.} Disabling deterministic result reordering raises process-runner peaks from 59\,k and 16\,k to 62.1\,k and 20.5\,k\,samples/s for text/VLM.
Further removing engine bookkeeping, lane execution, contributor tracking, and replay-control mechanisms brings them to 70.6\,k/20.6\,k\,samples/s. 
The guarantees thus cost 16\,\%/22\,\% relative to a stripped pipeline that can neither order its output nor checkpoint.}

\kmessage{\zephon's operator overhead is amortized by larger payloads.
On VLM data, it delivers $2.7\times$ higher throughput than the best baseline, while remaining sufficiently high for text data.}

\textbf{Multithreading.}
\zephon's thread runner on Python~3.13 is severely GIL-limited, plateauing \emph{below} the process runner despite avoiding IPC entirely.
The GIL serializes the fetch threads, preventing the runner from exploiting additional parallelism beyond the first few threads.
We evaluate \zephon{} on Python~3.14t, which removes the GIL (\Cref{fig:5-multithread}).
For text, the thread runner improves from 34\,k to  140\,k\,samples/s ($4.1\times$), now scaling up to DOP$=$16 rather than plateauing at DOP$=$4.
This brings it within reach of the best baseline (\textsc{Grain}/201\,k) while retaining \zephon's operator-graph flexibility.
The process runner also benefits modestly (59\,k${\to}$71\,k, $1.2\times$), as the pumps run concurrently with the Python main thread.

For VLM data, the effect is even more pronounced.  
The thread runner jumps from 10\,k to 59\,k\,samples/s ($5.9\times$), surpassing even the Python~3.14t process runner (27\,k) by $2.2\times$.
Where the process runner was the stronger VLM configuration on 3.13 (16\,k vs.\ 10\,k), GIL removal inverts this.

We do not report the worker-based data loaders, since their throughput only regresses due to the per-thread overhead in GIL-free Python.
So far, besides \zephon, only the \textsc{Spdl} project~\cite{Hira2025SPDL} is built for GIL-free data loading.
However, \textsc{Spdl} supports \emph{only} threads, whereas \zephon{} allows choosing between threads and processes as execution substrate while maintaining determinism.

\kmessage{Removing the GIL unlocks thread-based scaling.
On Python 3.14t, the thread runner closes the gap with workers on text and surpasses the process runner by $2.2\times$ on VLM data.}

\revi{\textbf{SHM coalescing.} We evaluate shared-memory coalescing in \zephon's process runner (\Cref{sec:execution:process-runner}) by disabling it (OS Pipe in~\Cref{fig:5-shm}).
For text data, this \emph{improves} throughput by 45\,\%, from 59\,k to 86\,k\,samples/s.
These samples are small, fixed-size NumPy arrays (${\sim}$16\,KiB each), so the overhead of shm allocation, copying, and metadata transfer exceeds the cost of direct serialization.}

\revi{For VLM data, the opposite holds: disabling coalescing collapses throughput from 16\,k to 2.4\,k\,samples/s ($-$85\,\%), and the process runner no longer scales.
VLM samples contain large, variable-size image buffers that quickly saturate the OS pipe buffer when serialized directly.
Thus, SHM coalescing amortizes IPC cost for large payloads, but its fixed overhead dominates when samples are small.}

\kmessage{SHM coalescing enables \zephon{} to achieve a $6.7\times$ throughput gain on VLM data over standard IPC.}

\subsubsection{VLM pipeline}\label{subsubsec:vlm-pipeline}

We benchmark a non-monotone, standard VLM preprocessing pipeline (\Cref{sec:background}).
After reading data from storage (\texttt{fetch}), \texttt{decode} parses text into conversations and decodes images to Pillow.
\texttt{Preprocess} validates image-text alignment, expands placeholders into vision-token sequences, applies the chat template, and runs image transforms like resize.
\texttt{Tokenize} converts the decorated text into tokens.
Afterwards, \texttt{shuffle+pack} performs first-fit bin packing.
For fairness, we share the same packing implementation across all loaders.
Last, \texttt{extract} prepares each packed sample for batching, and \texttt{construct} assembles a flat training batch compatible with FlashAttention-2~\cite{Dao2024FlashAttention2} and validates batch integrity.
Flat batches concatenate all documents into a single tensor without a batch dimension, so raw token throughput does not reflect capacity wasted by suboptimal packing.
We therefore report the normalized \emph{usable tokens per second} (utps), i.e., \revi{the total number of non-padding tokens} across all GPUs divided by wall-clock time and GPU count.
We normalize per-GPU to emphasize the scaling behavior of each loader.

\begin{figure}
  \includeinkscape[width=0.95\linewidth]{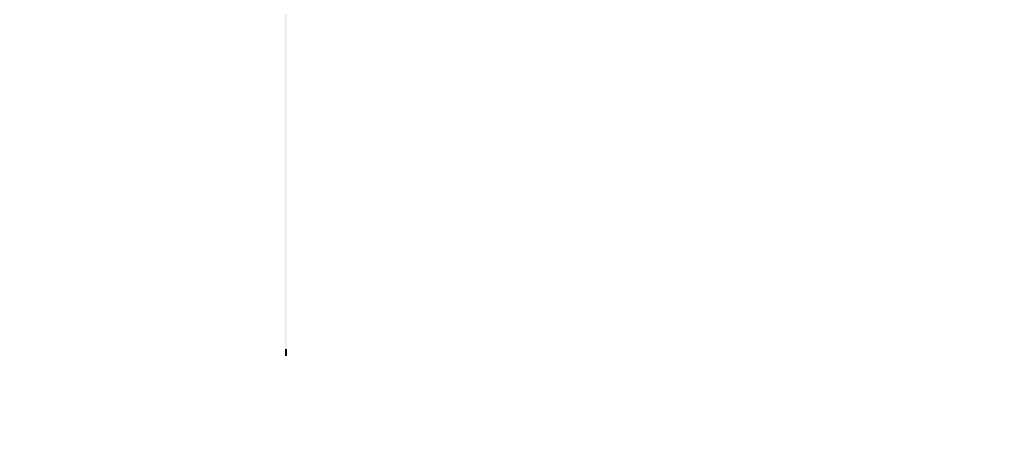}
  \caption{Per-GPU normalized VLM pipeline throughput depending on degree of parallelism.} %
  \label{fig:eval-vlm-scaling}
  \vspace{-0.5cm}
\end{figure}

\textbf{Loaders.} We compare \zephon{} to a \textsc{Grain} pipeline and to \textsc{InternalDs}, a solution built around \textsc{Streaming}~\cite{MosaicML2022Streaming} by (non-systems) ML researchers that runs all preprocessing in GIL-contending background threads in the main Python process.
We use \zephon{} with a fused implementation which merges \texttt{decode}, \texttt{preprocess}, and \texttt{tokenize} into a single operator, and use the process runner since the torchvision transforms are not thread-safe.
For \textsc{Grain} and \textsc{InternalDs}, the degree of parallelism (DOP) equals the number of worker processes.
In \zephon, aggregate DOP is the sum of per-operator parallelism across pipeline operators.
We sweep a few distributions of a fixed parallelism budget and report the best at each DOP.
\zephon's deterministic mode is enabled throughout.

\textbf{Results.} \Cref{fig:eval-vlm-scaling} shows the normalized usable token throughput as the DOP increases. 
\revi{\textsc{InternalDs}'s throughput plateaus as DOP increases because its GIL-contending threads cannot exploit additional parallelism.}
\revMRe{Its throughput also stays below the 16.9\,k\,utps per GPU required to avoid data stalls.}
\textsc{Grain} and \zephon{} both scale near-linearly.
At DOP\,32 their per-GPU throughput is effectively tied (29.8\,k vs.\ 29.9\,k).
At DOP\,64, \textsc{Grain} leads by ca. 8\,\% (26.6\,k vs.\ 28.7\,k). 
That \zephon{} matches \textsc{Grain} and is able to scale despite incurring IPC at each operator boundary is a direct consequence of our optimized SHM-coalesced transport (\Cref{sec:execution:process-runner}).
Bulk image and token data is placed in shared memory with only lightweight metadata crossing the pipe, and the flat payload layout minimizes serialization depth.
At DOP\,64, both systems deliver over $23\times$ the throughput of \textsc{InternalDs}.
Crucially, \zephon{} provides elastically deterministic output ordering across parallelism changes (\Cref{subsec:eval-elastic-determ}), which no other baseline provides.

\kmessage{\zephon{} matches \textsc{Grain}'s near-linear scaling on a VLM pipeline despite IPC and maintaining elastic determinism.}

\subsubsection{Time-to-first-batch}
\revi{Beyond steady-state throughput, startup time matters when iterating rapidly on short training runs.
We measure \emph{time-to-first-batch} (TTFB) from loader construction to the first batch being available.
We use the scenarios from~\Cref{subsubsec:fetchnbatch,subsubsec:vlm-pipeline} at DOP\,64.} %

\revi{\textbf{Process overhead.}
With the GIL, \zephon's thread runner has the lowest fetch-and-batch TTFB: 0.24/0.46\,s for text/VLM payloads, versus 2.87/3.03\,s with processes.
Spawning 64~worker processes dominates startup, whereas thread creation is near-instant.
Unlike for steady-state throughput, removing the GIL has little effect: TTFB falls to 0.22/0.43\,s with threads and 2.78/2.97\,s with processes.
\textsc{Streaming} (0.97/1.68\,s) and \textsc{Grain} (2.24/2.84\,s) fall between the two runners.
\textsc{LitData} is a striking outlier on text (17.92\,s).
Its TTFB scales superlinearly with worker count, as each worker independently initializes shard metadata.
For VLM payloads, fewer samples per shard reduce this overhead (2.27\,s).}

\revi{\textbf{VLM pipeline.}
The full VLM pipeline amplifies TTFB differences.
\textsc{InternalDs} requires 101.17\,s, as its GIL-contending background threads serialize all preprocessing during startup.
With the GIL, \zephon's process runner achieves 7.61\,s versus \textsc{Grain}'s 9.21\,s, reducing TTFB by 17\,\% relative to \textsc{Grain} and by a factor of $13$ relative to \textsc{InternalDs}.}

\kmessage{\zephon's thread runner delivers subsecond TTFB by avoiding process-spawn overhead.}

\subsection{Checkpointing analysis}\label{subsec:eval-checkpointing}

\revMRb{We evaluate the non-monotone VLM pipeline on eight parallel processes. S3 stores all data, aggregated states, and checkpoints; each run starts with an empty NVMe dataset cache.
We repeat each configuration three times per seed and report medians.
}

\textbf{In-flight state.} \revMRb{We examine how flush sentinel cadence $K$ and chunk size $C$ affect the retained state.
We sweep $K\in\{1,2,8,16,32\}$ and $C\in\{1\,024,2\,048,4\,096\}$ with three seeds, checkpointing after 100 simulated training steps.
Each step consumes four batches per rank without model computation.
With increasing $C$, each rank holds $26$/$13$/$7$ in-flight chunks, corresponding to $26.6$/$26.6$/$28.7$\,k in-flight records.
The nearly constant record count is caused by backpressure limiting read-ahead.
Changing the flush cadence $K$ has a negligible effect on the in-flight state.

To confirm how the state scales with training progress, we fix $K=8$, $C=1\,024$, and checkpoint after 1\,000/10\,000 steps.
The checkpoints retain an almost-constant amount of 29.8/30.3\,k in-flight records per rank (26.6\,k at step 100).}
\revMRd{Frequent flushes may have an impact on stream quality, such as packing efficiency.
However, we find that even in the worst case ($K=1$, $C=1\,024$) flushing increases token waste by less than $0.37$\,pp relative to no flushing.}

\textbf{Save cost.} \revMRb{In the  $K/C$ sweep (step 100), saving to S3, including aggregation, takes 3.03\,s (IQR 2.37--3.54\,s).
Checkpoint size grows only modestly with training step (2.61/2.92/2.97\,MiB for 100/1\,k/10\,k steps).
Saving takes 3.56\,s at both later checkpoints.
}

\textbf{Restore cost and replay overhead.} \revMRb{We measure TTFB from checkpoint reading to the first batch, including pipeline startup and replay.
At step~100, the median TTFB across the $K/C$ sweep is 48.5\,s, compared to 44.8\,s for a cold start (+8\,\%).
Restoring checkpoints taken at steps~1\,k/10\,k takes 70.2/73.8\,s.
}

\revMRb{\textbf{Lane scaling.} We test recovery with more than one lane per rank, fixing $K=8$, $C=1\,024$, and step~100.
We increase $L$ to 64 and 512 (8/64 lanes per rank).
In-flight records count per rank rises from 26.6\,k  to 35.1/131.1\,k; checkpoint sizes increase to 3.46/13.10\,MiB.
Saving takes 3.47/8.76\,s, and TTFB after restore is 62.6/403\,s.}

\kmessage{\revMRb{Checkpoint state grows only marginally with training progress; more lanes per rank increase in-flight state and recovery cost.}}

\begin{figure}
  \begin{subfigure}{0.569\linewidth}
    \includeinkscape[width=\linewidth]{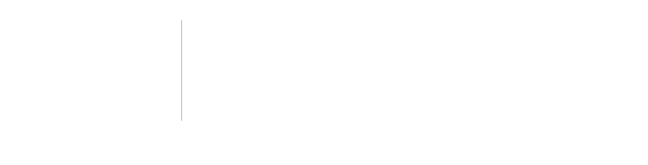}

      \captionsetup{margin={18pt,0pt}}
      \vspace{3.0ex} %
      \caption{Core v1}
      \label{fig:eval-core}
    \end{subfigure}%
    \begin{subfigure}{0.431\linewidth}
      \includeinkscape[width=\linewidth]{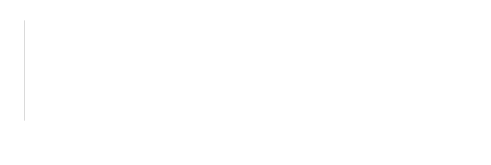}
      \vspace{0.2ex} %
      \caption{FineWeb}
      \label{fig:eval-fineweb}
    \end{subfigure}

    \par\vspace{-7.3ex}{\centering Spread across GPU counts (pp)\par}
    \vspace{2ex}
    
    \caption{\revMRf{Score spread (max - min) across GPU counts (8, 16, 32, 64) for 20\,B-token training.
    The dashed line is the reference level above which differences are meaningful.}}

    \label{fig:eval-scores}
    \Description{TODO!}
    \vspace{-0.5cm}
\end{figure}

\subsection{Elastic determinism}\label{subsec:eval-elastic-determ}

In~\Cref{fig:intro-el} we demonstrate how changing the GPU count leads to different loss curves. 
\revMRf{We now quantify the effect of this on downstream model accuracy.} 
\revi{Then, we evaluate \zephon's elastic determinism on a non-monotone VLM pipeline (\textbf{R3}).}

\revMRf{We reuse the architecture, hyperparameters, and seed checkpoint from~\Cref{fig:intro-el} and train a 1\,B model for 20\,B tokens (Chinchilla-optimal~\cite{Hoffmann2022ScalingLaws}) on 8, 16, 32, and 64 GPUs, both with and without elastic determinism. 
We then evaluate the final model of each run on the DCLM Core~v1~\cite{Li2024DCLM} and FineWeb~\cite{Penedo2024FineWeb} evaluation suites.
We repeat this for two seeds.}

\revMRf{\Cref{fig:eval-scores} shows the  max. spread of scores across GPU counts for both loaders, averaged across the two seeds.
DCLM  describes differences within $0.2$ Core~v1 points as comparable.
For FineWeb,  $\sim0.15$\,pp is the smallest uplift reconstructed from the reported results for the selected filters~\cite{Li2024DCLM,Penedo2024FineWeb}.
Without elastic determinism, the spread is $0.922$\,pp on Core~v1 and $0.555$\,pp on FineWeb, about four times the reference level.
With \zephon, however, the spread is just $0.061$\,pp and $0.007$\,pp, landing comfortably below the threshold. 
Using \zephon, downstream performance is thus determined by the run instead of by the GPU topology it happened to be trained on.}

\revMRf{\textbf{VLM.} To test \zephon{} in a second setting, }we train a 2.1\,B-parameter VLM consisting of a frozen SigLIP2-SO400M vision encoder~\cite{Tschannen2025SigLIP2}, a two-layer MLP projector, and a Qwen3-1.7B language backbone~\cite{Yang2025Qwen3}, for a total of ca. 1.7\,B trainable parameters.
We train for 300 steps with a global batch size of 256 using AdamW and a warmup-stable-decay learning rate schedule~\cite{Zhai2022LiT,Hu2024MiniCPM,Hagele2024WSD,Wen2025WSD} (20-step warmup, peak LR $5\cdot10^{-4}$).
\vldb{\revMRh{We do not train the model to convergence because this experiment is designed to demonstrate the effect of elastic determinism rather than final model quality.
The shorter run makes the loss curves easier to inspect in detail.}
}
\revi{To isolate the effect of the input sequence, we minimize numerical variation in the training framework. We train in fp32, use SDPA attention, enable all deterministic CUDA options, and initialize every run from the same seed checkpoint. With lower precision, loss curves can differ more across GPU configurations even when the global batches are identical.}
We compare \zephon{} and the \textsc{InternalDs} solution based on \textsc{Streaming} and train the model using 8, 16, 32, and 64 H100 GPUs.

\begin{figure}

    \begin{subfigure}[b]{0.556\linewidth} %
      \includeinkscape[width=\linewidth]{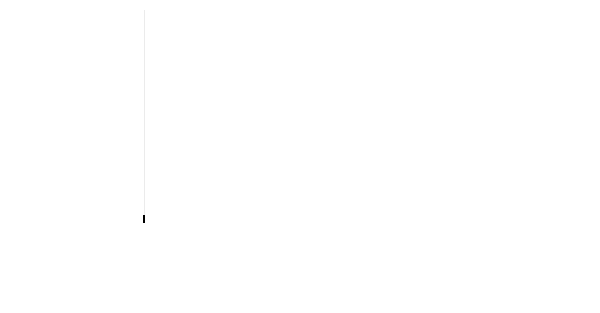}
    \captionsetup{margin={14pt,0pt}}
      \caption{\textsc{InternalDs}}
    \end{subfigure}%
    \begin{subfigure}[b]{0.444\linewidth} %
      \includeinkscape[width=\linewidth]{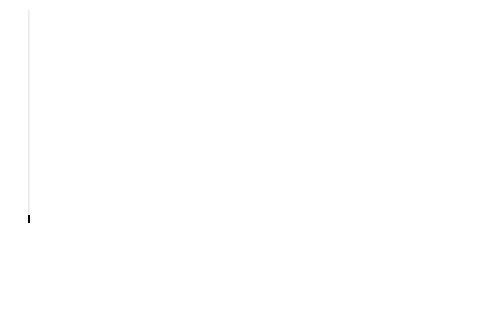}
      \caption{\zephon}
    \end{subfigure}
    \caption{VLM training loss across 8–64 GPUs.}
    \label{fig:eval-vlm-determinism}
    \vspace{-0.5cm}
  \end{figure}

The results are shown in~\Cref{fig:eval-vlm-determinism}.
Without elastic determinism, each GPU count produces a distinct loss trajectory.
\revi{\zephon{} produces identical global batches for every GPU count;  the corresponding loss curves are closely aligned, with \zephon's mean absolute pairwise difference being $5.84\times$ lower.} %

\kmessage{\revi{\zephon{} produces identical global batches across 8–64 GPUs on a stateful, non-monotone VLM pipeline.}}

\section{Related work}

\textbf{Foundation model data.} The ML community is exploring how to best filter~\cite{Gadre2024DataComp,Li2024DCLM,Penedo2024FineWeb}, mix~\cite{Xie2023DoReMi,Chen2023Skillit,Ye2025DataMixingLaws,Liu2025RegMix,Chen2026OLMix}, and curate and process~\cite{Penedo2024DataTrove,Soldaini2024Dolma,Huang2025DataGen,Maini2024WRAP,Maini2025BeyondWeb,Niklaus2026SyntheticPlaybook,NVIDIA2025Nemotron3Nano} training data for foundation models.

\textbf{Data loaders.}
Data stalls have motivated data loader research since the late 2010s~\cite{Yang2019AcceleratingDL,Zhu2018DeepIO,Mohan2021DataStalls,Isenko2022Bottleneck,Kuchnik2022Plumber,Bachkaniwala2024Lotus}.
\revi{Operator-centric loaders organize pipelines as operator graphs~\cite{Murray2021tfdata,Zhao2024cedar}. %
Format-centric loaders optimize specific data formats~\cite{MosaicML2022Streaming,Aizman2019WebDataset,Chaton2023LitData,Gruener2018Petastorm}.
There is work on disaggregated data loading~\cite{Audibert2023TFDataService,graur2022cachew,graur2024pecan,Um2023FastFlow,Zhang2025DistTrain} and sharing loaded data across \revi{co-located} jobs~\cite{Kumar2020Quiver,Xie2022Joader,Robroek2026TensorSocket}.}
\textsc{MinatoLoader}~\cite{Nouaji2026MinatoLoader} improves throughput by reordering samples, explicitly trading determinism for throughput.
Foundation model-specific systems include \textsc{Mixtera}~\cite{Bother2026Mixtera} and \textsc{MegaScale-Data}~\cite{Zhao2025OVERLORD}.
\revMRf{\textsc{Mosaic Streaming} introduces \emph{elastic determinism}, i.e., the preservation of the global sample order across changes in the number of GPUs, nodes, or CPU workers~\cite{MosaicMLStreamingElasticDeterminism}.
It implements this for indexable datasets by partitioning the sample space into canonical nodes, which are remapped to the physical topology.
\textsc{Grain}~\cite{Grain2023Github} similarly provides elastic determinism for indexable pipelines and deterministic resumption for stateful pipelines under fixed execution configurations.}
\revDc{Autotuning~\cite{graur2022cachew,graur2024pecan,Gavriilidis2025XDBC,Didrich2025Adapt} can adjust data loader operator parallelism.}

\revMRf{\textbf{Elastic training and recovery.}
\textsc{EasyScale}~\cite{Li2023EasyScale} preserves bitwise model state across GPU counts using logical workers and saved index/RNG state;
\textsc{Tenplex}~\cite{Wagenlander2024Tenplex} preserves sample order by repartitioning immutable, indexed data.
Both assume indexable datasets.
Complementarily, \textsc{Gemini}~\cite{Wang2023Gemini} accelerates checkpoint recovery under fixed resources, but leaves ordering out of scope.}

\textbf{Stream processing and dataflow systems.}
Checkpointing stateful pipelines over unbounded streams is a classic problem in stream processing~\cite{Akidau2013MillWheel,Carbone2015Flink}.
For example, \textsc{Flink} adapts  Chandy--Lamport distributed snapshots~\cite{Chandy1985Snapshots,Carbone2015ABS}.
\textsc{Naiad}~\cite{Murray2013Naiad} introduces timely dataflow and pointstamp-based progress tracking for cyclic, iterative dataflow graphs.
Data loading pipelines are structurally simple: each lane is a single-input operator chain with no joins or triggers.
\revMRf{While \zephon's operator toolbox is tailored for foundation model training, the system could be used for any stateful $n$-to-$m$ dataflow graph requiring reproducible ordered output.}

\textbf{Deterministic parallelism.}
Prior work explores deterministic parallelism at the language (DPJ~\cite{Bocchino2009DPJ}), compiler (\textsc{CoreDet}~\cite{Bergan2010CoreDet}, \textsc{Kendo}~\cite{Olszewski2009Kendo}), or OS level (\textsc{DThreads}~\cite{Liu2011DThreads}, \textsc{Determinator}~\cite{Aviram2010Determinator}).

\section{Conclusion}

\vldb{\revi{We present \zephon, to our knowledge, the first data loader to provide elastic determinism and efficient resumption for online, stateful $n$-to-$m$ foundation model data pipelines}} by serializing ordering decisions in accumulators, while parallelizing stateless work on interchangeable backends.
With \zephon, we can run more reliable ablations while reducing our compute-spend thanks to \zephon's higher and more stable throughput for VLM training and low startup time on trillion-token-scale text corpora.

\bibliographystyle{ACM-Reference-Format}
\bibliography{meta/bibliography}

\end{document}